# When Neurons Fail

El Mahdi El Mhamdi
*EPFL*
*Lausanne, Switzerland*
elmahdi.elmhamdi@epfl.ch

Rachid Guerraoui
*EPFL*
*Lausanne, Switzerland*
rachid.guerraoui@epfl.ch

*Abstract*—Neural networks have been traditionally considered robust in the sense that their precision degrades gracefully with the failure of neurons and can be compensated by additional learning phases. Nevertheless, critical applications for which neural networks are now appealing solutions, cannot afford any additional learning at run-time.

In this paper, we view a multilayer neural network as a distributed system of which neurons can fail independently, and we evaluate its robustness in the absence of any (recovery) learning phase. We give tight bounds on the number of neurons that can fail without harming the result of a computation. To determine our bounds, we leverage the fact that neural activation functions are Lipschitz-continuous. Our bound is on a quantity, we call the *Forward Error Propagation*, capturing how much error is propagated by a neural network when a given number of components is failing, computing this quantity only requires looking at the topology of the network, while experimentally assessing the robustness of a network requires the costly experiment of looking at all the possible inputs and testing all the possible configurations of the network corresponding to different failure situations, facing a discouraging combinatorial explosion.

We distinguish the case of neurons that can fail and stop their activity (*crashed* neurons) from the case of neurons that can fail by transmitting arbitrary values (*Byzantine* neurons). In the crash case, our bound involves the number of neurons per layer, the Lipschitz constant of the neural activation function, the number of failing neurons, the synaptic weights and the depth of the layer where the failure occurred. In the case of Byzantine failures, our bound involves, in addition, the synaptic transmission capacity. Interestingly, as we show in the paper, our bound can easily be extended to the case where synapses can fail.

We present three applications of our results. The first is a quantification of the effect of memory cost reduction on the accuracy of a neural network. The second is a quantification of the amount of information any neuron needs from its preceding layer, enabling thereby a boosting scheme that prevents neurons from waiting for unnecessary signals. Our third application is a quantification of the trade-off between neural networks robustness and learning cost.

## I. Introduction

Since their inception in the 1940s [1], artificial neural networks received an oscillating amount of interest. They went trough two periods of excitement, each followed by a loss of interest, before the current popularity. After the very first period of "discovery", from the 1950s [2] to the late 1960s, came the *AI winter*, when Minksy questioned the ability of perceptrons to learn non linearly separable functions such as the exclusive OR [3]. Neural networks received a regain of interest in the late 1970s with the back-propagation algorithm [4] that overcame learning issues. Interest vanished again in the 1990s due to the lack of computing power, even-tough important practical achievements were done in the late 1980s and early 1990s in the field of image recognition [5].

Neural networks are now back in the headlines for their outstanding performance [6] in tasks such as function approximation, image and speech recognition, as well as weather prediction [7]. Today, neural networks are considered in far more critical applications than those of the last decade: flight control [8], radars [9] and electric cars [10]. This motivates a better understanding of the extent to which neural networks can be *robust*.

A simple approach to achieve robustness is to consider the entire neural network as a single piece of software [11], replicate this piece on several machines, and use classical state machine replication schemes to enforce the consistency of the replicas [12]. In this context, no neuron is supposed to fail independently: the unit of failure is the entire machine hosting the network. However, forcing an entire network to run on a single machine clearly hampers scalability. One could also consider strict subsets of the neural network as different pieces of software, each running on one Turing machine [13]. In this case, classical replication schemes can still be applied, but one has to face usual distributed computing problems, e.g., to handle the synchronization of message passing between subsets of the network [14].

Biological plausibility, together with scalability, call for going one step further and considering each neuron as a *single* physical entity (that can fail *independently*), i.e., to go for genuinely distributed neural networks [15]. This approach is considered for example in the Human Brain project [16], trying to emulate the mammalian brain or the various works on hardware-based neural networks [17]. More recently, teams from IBM reported [18], [19] a successful neuromorphic implementations of convolutional neural networks that require a running power as low as 25 mW to 275 mW. In those settings, the unit of failure is one single neuron or synapse, and not a whole machine.

In this paper, we explore this granularity and view a neural network as a distributed system where neurons can fail independently. We ask what is the maximum number of such failures that can be masked by the neural network, i.e., without having any impact on the overall computation. Addressing this question goes, however, first through precising it.

It is actually well known that the failure of neurons can be tolerated through additional learning phases [20], [21]. However, stopping a neural network and recovering its failures through a new learning phase is not an option for critical applications [8], [10], [9]. One can also consider specific a priori learning schemes that make it possible to tolerate failures a posteriori, e.g., shutting down parts of the network while learning, in order to cope with failures at run-time (dropout) [6], [22].

Our question can then be posed as follows: if (a) we do not make any assumption on the learning scheme and (b) we preclude the possibility of adding learning phases to recover from failures when the neural network is in the deployment phase, what is the maximum number of faulty neurons that can be tolerated?

At this point, the question might sound trivial and the answer could be simply: *none*. Indeed, how could a neural network tolerate failures if it was not specifically devised with that purpose in mind? More specifically, if the failures of a number of neurons do not impact the overall result, then these neurons could have been eliminated from the design of that network in the first place. In fact, the reason why the question is nontrivial is *over-provisioning* [23]. Indeed, neural networks are rarely built with the minimal number of neurons to perform a computation. To estimate exactly this minimal number, one needs to know the target function the network should approximate, which by definition is unknown since the sole mainspring for machine learning is that we only know a finite number of the values of the target function: the learning data set. In fact, it has been experimentally observed that over-provisioning leads to robustness [24], [25], [26], [22]. Yet, the exact relation between the over-provision and the actual number of failures to be tolerated has never been precisely established. This paper establishes this relation for the first time.

More precisely, we present tight bounds on the number of faulty neurons a *feed-forward neural network*[1] can tolerate without harming the computation result, nor requiring any additional learning phase or specific prior learning algorithm. In fact, our bounds are not simply expressed in terms of *numbers of failures*, but in terms of *weight* and *failure* distribution. Indeed, unlike process failures in traditional distributed computing that all have the same effect, neuron failures do not: they are *weighted*. In the general case of *multilayer*, or so called *deep*, networks, we formulate our results in the form of a *fault-per-layer distribution*.

Our results are obtained using analytic properties of the different mathematical components of a neural network, namely the activation function, the synaptic weights, and the neural computation model. By relying on the very fact that neural activation functions are bounded, and in practice Lipschitzian [27], we set precise bounds on the error propagation over the layers, and subsequently establish tight bounds on the number of failures a neural network can tolerate.

For didactic reasons, we present our results in an incremental manner. We consider first a network with a single layer and focus on the crashes of neurons, then we generalize to a multilayer network with Byzantine (arbitrary [28]) failures of neurons. We show that if the transmission capacity of synapses is unlimited, no neural network can tolerate the presence of a single Byzantine neuron. Inspired by results from biophysics [29] and neuroscience [30], we consider, however, synapses with a limited transmission capacity, and give a bound on the Byzantine failures of neurons as a function of this capacity. Finally we show how bounds on synapses failures can be derived from bounds on neurons failures.

We discuss several applications of our results. The first is a bound on the effect on output accuracy of reducing the computational precision per neuron. This effect has been recently highlighted experimentally [31], however, without any theoretical explanation until the present. The second is a synchronization scheme that reduces the waiting time for neurons. The third is a quantification of the trade-off between robustness and ease of learning. We also discuss how to adapt our bounds to convolutional networks.

The rest of the paper is organized as follows. In Section II we model a neural network as a distributed system and state the robustness criteria. In Section III we prove an upper bound on crash failures in the case of single-layer neural network. We use this proof as a starting point for the generalization to Byzantine failures and given in Section IV, first for neurons and then for synapses in the multilayer case. We discuss in Section V some applications of our results. We conclude the paper by discussing other models and future work.

## II. MODEL

### A. Viewing a Neural Network as a Distributed System

We distinguish two main kinds of components of a neural network.

**Neurons.** These are the computing nodes (processes) of a neural network. They are either correct, in which case they execute their assigned computation (see below), or they fail, in which case they can stop computing (crash) or even send an arbitrary value (Byzantine faults). The failure of any neuron is independent from the failure of any other.

---
[1]Feed-forward networks are the most common in the literature [26], and today's most popular topology: the convolutional neural network for example [5], is a particular case of the feed-forward topology. We will discuss some of these in Section V.

Neurons communicate via message-passing [14] through synchronous point-to-point communication channels called *synapses*.

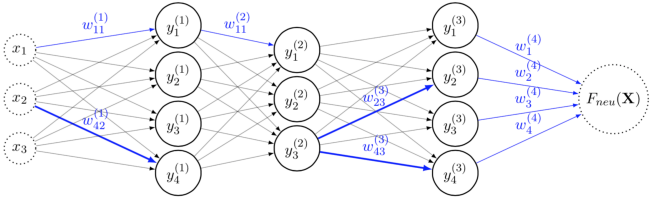

Figure 1. A (feed forward) neural network (solid nodes and edges), with $d = 3$, $L = 3$, $N_2 = 3$ and $N_1 = N_3 = 4$. Input and output nodes (dotted) are not considered as parts of the network, but as its clients. For readability, only some synaptic weights are represented (bold blue). $\mathbf{X} = (x_1, \ldots, x_d)$.

**Synapses.** These are the communication channels connecting the neurons. Just like neurons, synapses are either correct, in which case they transmit the signal provided to them (see below) or fail, in which case they stop transmitting the signals, or they transmit arbitrary signals. The failure of a synapse is also independent from that of other synapses and neurons. Synapses are weighted[2]. The weight models the importance a neuron $j$ gives to the signals emitted by a neuron $i$ at the other end of the synapse and is therefore also called the weight from neuron $i$ to neuron $j$. Faults at synapses can then be modeled as errors in the value of the weight: a *crashed synapse* is viewed as weighted by value 0 (stops transmitting), whereas a *Byzantine synapse* transmits any other value than the nominal value it is supposed to transmit, within its *capacity*.

Indeed, synapses have a *bounded transmission capacity*. This assumption is supported by two important works in biophysics [29] and neuroscience [30]. Hence if a faulty neuron corrupts the value it is supposed to send, the transmitted value is limited by the highest amount of electric charge flow the synapse can transport to the next neuron.

**Assumption 1.** *(Bounded transmission) There exists an upper bound $C \in \mathbb{R}^{*+}$ such that, for any input and any Byzantine neuron, the value transmitted by any synapse from that Byzantine neuron is bounded by $C$ in absolute value.*

When Assumption 1 is not satisfied, we say that the network has *unbounded transmission*.

**Neural Computation.** The effectiveness of feed-forward neural networks relies on a fundamental theorem [32] that guarantees their universal approximating power with as few[3] as one single layer.

Let $\epsilon$ be any positive real number (an accuracy level), and $F$ any continuous function mapping $[0,1]^d$ to $[0,1]$. The goal is to build an approximation of $F$ with accuracy $\epsilon$ (as constructed in the classical model of a multilayer perceptron [26]) which we abstract in the following description:

Neurons are distributed over a series of layers. We denote by $L$ the number of layers, each identified with index $l$ and containing $N_l$ neurons. Any neuron of layer $l-1$ is said to be *on the left* of any neuron of layer $l$ (layer $l$ is *on the right* of layer $l-1$). Each neuron *fires* (broadcasts) a signal (message) to all the neurons of the layer on its right. Neuron $j$ at layer $l$ receives the sum given by $s_j^{(l)}$ of Equation 1, where $y_i^{(l-1)}$ and $w_{ji}^{(l)}$ denote respectively the output value at neuron $i$ of layer $l-1$, and the weight of the synapse from that same neuron to neuron $j$ of the next layer $l$. To define its own output $y_j^{(l)}$, neuron $j$ of layer $l$ in turn injects the sum given by Equation 3 into a non-linear activation function, called a *squashing* function, $\varphi$, after adding a bias[4].

$$F_{neu}(\mathbf{X}) = \sum_{i=1}^{N_L} w_i^{(L+1)} y_i^{(L)}(\mathbf{X}) \quad (1)$$

$$\text{with } y_j^{(l)} = \varphi(s_j^{(l)})(l \geq 1); \; y_j^{(0)}(\mathbf{X}) = x_j \quad (2)$$

$$\text{and } s_j^{(l)} = \sum_{i=1}^{N_{l-1}} w_{ji}^{(l)} y_i^{(l-1)} \quad (3)$$

**Definition 1.** *(Approximation) We denote by $A = C([0,1]^d, [0,1])$ the space of continuous functions mapping $[0,1]^d$ to $[0,1]$. $F_{neu}$ as defined by Equation 1 is said to be a neural $\epsilon$-approximation of a target function $F \in A$ if we have: $\forall \mathbf{X} \in [0,1]^d$: $\|F(\mathbf{X}) - F_{neu}(\mathbf{X})\| \leq \epsilon$.*

**Universality.** We recall the universality theorem for a single layer network[5]: Let $d$ be any integer and $\varphi : \mathbb{R} \to [0,1]$ a strictly-increasing continuous function, such that $\lim_{x \to -\infty} \varphi(x) = 0$ and $\lim_{x \to +\infty} \varphi(x) = 1$. Given any function $F \in A$ and $\epsilon > 0$, there exist an integer $N(\epsilon)$, and a set of coefficients $(w_{ji}^{(1)})_{1 \leq j \leq N(\epsilon)}^{1 \leq i \leq d}$ and $(w_i^{(2)})_{1 \leq i \leq N(\epsilon)}$ such that $F_{neu}$ defined in Equation 1 is a neural $\epsilon$-approximation of $F$.

**Activation Function.** This function, denoted $\varphi$ is the essence of the non-linearity of neural networks. The universality theorem holds for any non-constant, bounded and monotonically increasing activation function $\varphi$. Yet, two main popular choices for $\varphi$ in machine learning applications are the logistic function *sigmoid* given by: $sigmoid(x) = \frac{1}{1+e^{-x}}$ and the hyperbolic tangent *tanh*.

In this paper, we only impose on $\varphi$ the conditions of the universality theorem, and consider that $\varphi$ is K-*Lipschitzian*,

---

[2] The weights are determined by the initial learning phase, when training the network.

[3] Note that universality for L=1 is harder to obtain than for $L > 1$: fewer layers to approximate the target function.

[4] Following the usual notational convenience [26], we omit the bias in the computation model. This is done without loss of generality, considering an additional constant neuron (value = 1) in each layer. During the learning phase – when building the network – instead of learning its bias value, the neuron of layer $l$ just learns the weight given to the constant neuron of layer $l-1$. As this weight is always multiplied by 1, the weight serves as the bias, around which the activation function is be centered.

[5] The interested reader can refer to the proof of [32].

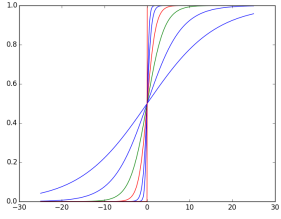

Figure 2. The profile of a sigmoid function, centered around 0 and tuned with several values of K. The larger is K, the steeper is the slope and the more discriminating is the activation function at each neuron.

meaning that $K = \sup |\frac{\varphi(x)-\varphi(y)}{x-y}|_{x \neq y}$ exists and is a finite real number.

Moreover, $K$ can be tuned. Consider for instance the commonly used function, *sigmoid*, this function is $\frac{1}{4}$-Lipschitzian but can be tuned to be K-*Lipschitzian* (Figure 2), by taking in this case $x \mapsto \varphi(4Kx)$ as the K-tuned activation function. (The detailed derivation of the Lipschizness of $\varphi$ is given in a companion technical report [33].). In this paper we consider, without loss of generality, *sigmoid* as the choice for $\varphi$.

### B. Failures and Robustness

**Definition 2.** *(Failures) We say that a neuron $i$ in layer $l$ crashes when neuron $i$ stops sending values, in which case $y_i^{(l)}$ is considered[6] to be equal to 0 by other neurons[7]. We say that neuron $i$ is Byzantine, when $y_i^{(l)}$ is arbitrary.*

**Definition 3.** *(Robustness) We say that a neural $\epsilon$-approximation $F_{neu}$ of a target function $F$ realized by $N$ neurons tolerates $N_{fail}$ faulty neurons, if for any subset of neurons $I_{fail} \subset \{1, \cdots, N\}$ of size $N_{fail}$, we can modify $F_{neu}$ for the failing neurons according to Definition 2 and still $\epsilon$-approximate $F$ by $F_{neu}$.*

**Lemma 1.** *With unbounded transmission, no neural network can tolerate a single Byzantine neuron.*

*Proof:* As a consequence of the neural computation and definitions 2 and 3, if the transmission is unbounded, a Byzantine neuron at layer $L$ sending a value higher than $\epsilon$ plus the difference between the nominal $F_{neu}$ and the contribution of the remaining neurons breaks the $\epsilon$-approximation as stated in Definition 1. ∎

### C. Over-Provisioning

Using the universality theorem, we can define a minimal number of neurons $N_{min}(\epsilon)$ below which the neural network cannot yield an $\epsilon$-approximation of $F$. By definition of $N_{min}(\epsilon)$, if a neural network is built with $N_{min}(\epsilon)$ neurons, the network cannot tolerate any crashed neuron.

Clearly, neural networks are not robust, they do not tolerate any neuron failure when built with the minimal amount of neurons. But, as we discussed in the introduction, this is usually not the case: they are over-provisioned [23] and contain more than $N_{min}(\epsilon)$ neurons. With the work of Barron [34], we know that $N_{min}(\epsilon) = \Theta(\frac{1}{\epsilon})$ and that given $N$ neurons, a network can achieve an error in the order of $\frac{1}{N}$ when $N$ is large. Instead of looking at over-provisioned networks as containing more than $N_{min}(\epsilon)$, we consider the quality of the approximation they are providing: given $\epsilon' \leq \epsilon$ and $F_{neu}$ a neural $\epsilon'$-approximation of $F$, $F_{neu}$ is also a neural $\epsilon$-approximation of $F$ and is said to be an over-provisioned $\epsilon$-approximation of $F$.

In the following, we set the conditions under which a neural network, realizing an $\epsilon'$-approximation ($\epsilon' \leq \epsilon$) of $F$, can tolerate $N_{fail}$ failures and keep realizing an $\epsilon$-approximation of $F$. For convenience, all the bounds on the failures are stated in terms of $\epsilon$ and $\epsilon'$.

## III. SINGLE-LAYER NEURAL NETWORKS

For didactic reasons, we first start with the case of a single layer neural network. We translate the fact that the network tolerates the crash of $N_{fail}$ neurons as given by Definition 3 to an inequation, which we combine with an estimation of the distance between the value of the damaged network and the nominal value of the output (which we recall is close to the target by a distance $\epsilon'$). We end up with an upper bound on $N_{fail}$.

To prove that the bound is tight, we look at the worst failure case. Intuitively, this corresponds, following the tradition in distributed computing [14], to an adversary killing "key neurons": those with highest weights, and looking at an input were those same neurons were instrumental: broadcasting the highest possible value $y_j^{(1)}$, as close to 1 as possible.

**Theorem 1.** *Let $F$ be any function mapping $[0,1]^d$ to $[0,1]$. Let $\epsilon$ and $\epsilon'$ be any two positive real numbers such that $0 < \epsilon' \leq \epsilon$. For any neural $\epsilon'$-approximation $F_{neu}$ of $F$ (Definition 1) and any integer $N_{fail}$: If $N_{fail} \leq \frac{\epsilon-\epsilon'}{w_m}$ where $w_m = max(\|w_i^{(2)}\|, i \in [1, N])$ is the maximum norm of a weight from the single layer to the output node, then $F_{neu}$ is a neural $\epsilon$-approximation of $F$ that tolerates $N_{fail}$ crashed neurons (Definition 3). The bound on $N_{fail}$ is tight.*

*Proof:* **Upper bound.** Applying the universality theorem[8] on $F$ and $\epsilon'$, let $F_{neu}$ be a neural $\epsilon'$-approximation of $F$ with $N$ the number of neurons of $F_{neu}$ and $w_m$ the maximal weight from the single layer of $F_{neu}$ to the output.

Let $N_{fail}$ be any integer such that $N_{fail} \leq \frac{\epsilon-\epsilon'}{w_m}$. Denote by $F_{fail}$ any of the modified values of the neural function $F_{neu}$ after $N_{fail}$ neurons crash: $F_{fail} = \sum_{i=1, i \notin I_{fail}}^{N} w_i^{(2)} y_i$, where $I_{fail}$ is a a set containing $N_{fail}$ crashed neurons. Let $\mathbf{X} \in [0,1]^d$ be any input vector. By the triangle inequality:

$$\|F(\mathbf{X}) - F_{fail}(\mathbf{X})\| \leq \|F(\mathbf{X}) - F_{neu}(\mathbf{X})\| \\ + \|F_{neu}(\mathbf{X}) - F_{fail}(\mathbf{X})\|. \quad (4)$$

---

[6] The strictly-increasing activation function $\varphi$ does not allow a correct neuron to output value 0.

[7] Remember that we assume synchronous transmission.

[8] The existence of a neural approximation for a given target function is taken here as granted by the universality theorem. One might wonder how do neurons, viewed as distributed processes, build the network (i.e put the correct weights to their linking synapses) in the first place to approximate that target function. This is done, during the learning phase, via the back-propagation algorithm [4]: neurons communicate in the reverse direction (from the output to the input) and re-adjust the weights *locally* according to the error value they are given by the output client.

Since $F_{neu}$ is an $\epsilon'$-approximation of $F$ we have:

$$\|F(\mathbf{X}) - F_{neu}(\mathbf{X})\| \leq \epsilon'. \quad (5)$$

From the definition of $F_{fail}$, we have: $\|F_{neu}(X) - F_{fail}(\mathbf{X})\| = \|\sum_{i=1, i \in I_{fail}}^{N} w_i^{(2)} y_i(\mathbf{X})\|$. Using another triangle inequality on norms we get:

$$\|F_{neu}(\mathbf{X}) - F_{fail}(\mathbf{X})\| \leq \sum_{i=1, i \in I_{fail}}^{N} \|w_i^{(2)}\| y_i(\mathbf{X}). \quad (6)$$

By definition of $w_m$ and the hypothesis on the bounded activation function, $\|w_i^{(2)}\| \leq w_m$ and $y_i(\mathbf{X}) \leq 1$ for all $\mathbf{X}$ and $i$. Inequality 6 becomes:

$$\|F_{neu}(\mathbf{X}) - F_{fail}(\mathbf{X})\| \leq \sum_{i=1, i \in I_{fail}}^{N} w_m = N_{fail} w_m \quad (7)$$

Merging inequalities 4, 5 and 7 we obtain: $\|F(\mathbf{X}) - F_{fail}(\mathbf{X})\| \leq N_{fail}.w_m + \epsilon'$.

Since $N_{fail} \leq \frac{\epsilon - \epsilon'}{w_m}$, we have $\|F(\mathbf{X}) - F_{fail}(\mathbf{X})\| \leq \epsilon$.

Therefore the upper bound on $N_{fail}$ : $N_{fail} \leq \frac{\epsilon - \epsilon'}{w_m}$, guarantees that $F_{fail}$, the neural function obtained from $F_{neu}$ with $N_{fail}$ crashed neurons is still an $\epsilon$-approximation of $F$.

**Tightness.** Let $N_{fail}$ be any integer such that $N_{fail} > \frac{\epsilon - \epsilon'}{w_m}$ and assume that $F_{neu}$ tolerates the crash of $N_{fail}$. Let $\delta = N_{fail} - \frac{\epsilon - \epsilon'}{w_m}$, by the initial assumption, $\delta > 0$.

Consider $\epsilon'$ to be the supremum on the approximation with which the over-provisioned neural network $F_{neu}$ approximates $F$, i.e $\epsilon' = sup_{\mathbf{X} \in [0,1]^d}(\|F(\mathbf{X}) - F_{neu}(\mathbf{X})\|)$.

Consider the equality cases as well as the limit cases (close to equality) for the key inequalities that lead to the upper bound on $N_{fail}$: In 4, equality occurs iff $F(\mathbf{X}) - F_{neu}(\mathbf{X})$ and $F_{neu}(\mathbf{X}) - F_{fail}(\mathbf{X})$ are positively proportional (equality case of the triangle inequality). In 6, equality occurs iff the weights of the crashed neurons are positively proportional. Assume an input and choice of crashed neurons satisfying both of these equality cases.

Let $\alpha$ be any positive real number. To be close to the limit case of Inequality 5, we can chose inputs $\mathbf{X}$ such that $\|F(X) - F_{neu}(X)\| > \epsilon' - \frac{\alpha}{2}$, those inputs exist otherwise $\epsilon'$ is not the supremum error achieved by $F_{neu}$ or $F$ is not a continuous function and we will have a contradiction.

In 7, the limit case corresponds to crashed neurons being those with maximal weights and inputs such that the neurons in $I_{crash}$ all output a value close to 1: let $\mathbf{X}$ be an input satisfying the previous equality and limit cases such that for any neuron $i$ in $I_{fail}$, we have $y_i(\mathbf{X}) > max(1 - \frac{\alpha}{2}, 1 - \frac{\alpha}{2(\epsilon - \epsilon')})$, i.e $y_i(\mathbf{X})$ close to 1. With this worst-case choice of input and crashed neurons, we obtain: $\|F(\mathbf{X}) - F_{fail}(\mathbf{X})\| = \|F(X) - F_{neu}(X)\| + \|\sum_{i=1, i \in I_{fail}}^{N} w_i^{(2)} y_i(\mathbf{X})\| > \epsilon' - \frac{\alpha}{2} + max(1 - \frac{\alpha}{2}, 1 - \frac{\alpha}{2(\epsilon - \epsilon')}) N_{fail}.w_m$.

Thus, in case of more crashes than allowed by the upper bound of the theorem ($N_{fail} > \frac{\epsilon - \epsilon'}{w_m}$) leads to:
$\|F(\mathbf{X}) - F_{fail}(\mathbf{X})\| > \epsilon' - \frac{\alpha}{2} + max(1 - \frac{\alpha}{2}, 1 - \frac{\alpha}{2(\epsilon - \epsilon')})(\epsilon - \epsilon' + \delta w_m) = \epsilon - \frac{\alpha}{2} + max(-\frac{\alpha}{2}, -\frac{\alpha}{2(\epsilon - \epsilon')})(\epsilon - \epsilon' + \delta w_m) + \delta w_m = \epsilon - \frac{\alpha}{2} - min(\frac{\alpha}{2}, \frac{\alpha}{2(\epsilon - \epsilon')})(\epsilon - \epsilon' + \delta w_m) + \delta w_m$.

If $\epsilon - \epsilon' \geq 1$ then $min(\frac{\alpha}{2}, \frac{\alpha}{2(\epsilon - \epsilon')}) = \frac{\alpha}{2(\epsilon - \epsilon')}$) and the latter inequality leads to $\|F(\mathbf{X}) - F_{fail}(\mathbf{X})\| > \epsilon - \frac{\alpha}{2} - \frac{\alpha}{2(\epsilon - \epsilon')}(\epsilon - \epsilon' + \delta w_m) + \delta w_m = \epsilon - \alpha + \delta'$, where $\delta' = \delta w_m(1 - \frac{\alpha}{2(\epsilon - \epsilon')}) > 0$ (for small $\alpha$).

If $\epsilon - \epsilon' < 1$ then $min(\frac{\alpha}{2}, \frac{\alpha}{2(\epsilon - \epsilon')}) = \frac{\alpha}{2}$ and the latter inequality leads to $\|F(\mathbf{X}) - F_{fail}(\mathbf{X})\| > \epsilon - \frac{\alpha}{2} - \frac{\alpha}{2}(\epsilon - \epsilon' + \delta w_m) + \delta w_m = \epsilon - \frac{\alpha}{2}(1 + (\epsilon - \epsilon' + \delta w_m)) + \delta w_m$ and since $\epsilon - \epsilon' + \delta w_m > 0$ this implies that $\|F(\mathbf{X}) - F_{fail}(\mathbf{X})\| > \epsilon - \frac{\alpha}{2} + \delta w_m$, which also leads to:
$\|F(\mathbf{X}) - F_{fail}(\mathbf{X})\| > \epsilon - \alpha + \delta'$ ($\delta' < \delta w_m$ for small $\alpha$ and $-\alpha < -\frac{\alpha}{2}$) since $\alpha > 0$).

In the inequality $\|F(\mathbf{X}) - F_{fail}(\mathbf{X})\| > \epsilon - \alpha + \delta'$.

$\alpha$ is any positive real number, for which we can chose an input leading to the inequality. Since $F$ and $F_{fail}$ are continuous, we can take the previous inequality to the limit $\alpha \mapsto 0$ and we abtain an inequality that contradict the assumption that $F_{neu}$ tolerates the crash of $N_{fail}$ neurons.

Therefore, by contradiction, the bound is tight. ∎

## IV. MULTILAYER NETWORKS AND BYZANTINE FAILURES

This section generalizes Theorem 1. While that theorem says that we can derive a tight bound on how many neurons can crash without losing $\epsilon$-accuracy, it does not capture the situation where neurons can send values different from those expected, whether this difference is arbitrary or controlled. The latter situation is that of correct neurons in a multilayer network: if a correct neuron has faulty neurons on its left[9], the output value of this neuron embeds some imprecision. The aim is to evaluate how the loss of accuracy propagates through layers and bound it on the output.

### A. Forward Error Propagation

Theorem 2 below says that, when errors occur at $f_l$ neurons of layer $l$, the effect is transmitted by all correct neurons at any layer $l'$ between layer $l$ and the output. This leads, in the worst case, to a series of multiplications, as many times as there are layers on the right before reaching the output, i.e $(L-l)$ times, by the Lipschitz constant, by the number of correct neurons at layer $l'$, i.e $(N_{l'} - f'_l)$, by the maximum weight $w_m^{(l')}$, by $f_l$ and by the bound $C$ of Assumption 1. The previous products are summed over the layers. As a calculation convention for the rest of the paper, we consider an $(L+1)$-th layer consisting of the output node with $N_{L+1} = 1$ correct neuron and $f_{L+1} = 0$ failing neurons (though it is not part of the neural network, unlike the $(L+1)$-th sets of synapses which are part of the network). Finally, the effect of a failure on the output increases exponentially with the depth of the layer (dependency on $K^{L-l}$).

---
[9]The conventions of *left* and *right* are defined in the neural computation described in Section II-A.

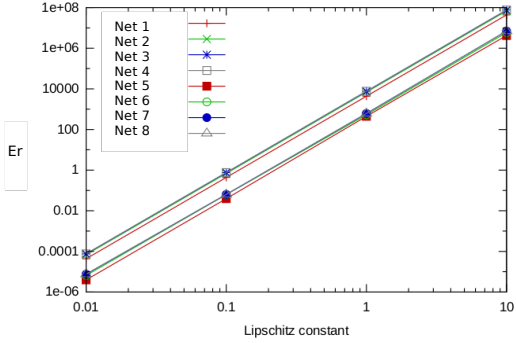

Figure 3. Experimental values of the error (Er) at the output of several neural networks, affected with similar amount of neuron failures, plotted against the Lipschitz constant in a log scale.

Denote by $Fep$ the quantity described above and given by the following equation:

$$Fep = C \sum_{l=1}^{L} \left( f_l K^{L-l} w_m^{(L+1)} \prod_{l'=l+1}^{L} (N_{l'} - f_{l'}) w_m^{(l')} \right).$$

Note that $Fep$ has a polynomial dependency on $K$ as observed in Figure 3.

**Theorem 2.** *Consider a neural network containing L layers. If in each layer $l$, $f_l$ neurons, among the $N_l$ neurons, are affected by errors such that any neuron $j$ within layer $l$ broadcasts an output $y_j^{(l)} + \lambda_j^l$ to the next layer instead of the nominal $y_j^{(l)}$, then the effect on the output is bounded as follows:*

$$\|F_{neu}(\mathbf{X}) - F_\lambda(\mathbf{X})\| \leq Fep \qquad (8)$$

*where $F_{neu}$ is the nominal neural function, $F_\lambda$ the neural function accounting for the errors $\lambda_j^{(l)}$, and $w_m^{(l)} = max(|w_{ji}^{(l)}|, (j,i) \in [1, N_l][1, N_{l-1}])$ is the maximum norm of the weights of the incoming synapses to layer $l$. The bound (8) is tight.*

*Proof:* We proceed by induction on $L$.

**Initiation.**

Let $N_{fail} = f_1$ be the number of neurons failing in the single layer of the network, let $I_{fail}$ be the set containing those neurons, we have:

$$\|F_{neu}(\mathbf{X}) - F_\lambda(\mathbf{X})\| = \|\sum_{i \in I_{fail}} w_i^{(2)}(y_i^{(1)} + \lambda_i^{(1)})\|$$

Which, by the triangle inequality leads to:

$\|F_{neu}(\mathbf{X}) - F_\lambda(\mathbf{X})\| \leq \sum_{i \in I_{fail}} \|w_i^{(2)}(y_i^{(1)} + \lambda_i^{(1)})\|$, equality cases occur for positively proportional terms (**Condition 1**). Applying Assumption 1 and the definition of $w_m(2)$ gives us:

$$\|F_{neu}(\mathbf{X}) - F_\lambda(\mathbf{X})\| \leq f_1 w_m^{(2)} C \qquad (9)$$

Equality cases occur for inputs such that $y_i^{(1)} + \lambda_i^{(1)} = C$ (**Condition 2**) and when the failing neurons are all linked to the output with the maximal weight $w_m^{(2)}$ (**Condition 3**).

We observe that Inequation 9 is the ($L = 1$) version of Theorem 2, and that similarly, due to the worst case of failures (i.e when Conditions 1 to 3 are simultaneously occurring), the bound is tight.

**Induction step.**

Assume Theorem 2 holds for networks with up to some number of layers $L \geq 1$.

Now consider a network consisting of $(L+1)$ layers. The layered structure of the network enables us to see each of the $N_{L+1}$ neurons of the $(L+1)^{th}$ layer, first as an output to an $L$-layer network (all the nodes to the left of that neuron), and second, after applying the activation function, as a neuron in a single-layer neural network (consisting of the $(L+1)^{th}$ layer alone).

In this last $(L+1)^{th}$ layer, we can distinguish two subsets of neurons:

1) (Failing neurons at layer L+1) A subset of $f_{L+1}$ failing neurons, that yields, as in the initiation step (sigle layer), an error of at most $f_{L+1} w_m^{(L+2)} C$.
2) (Correct neurons at layer L+1) A subset of $N_{L+1} - f_{L+1}$ correct neurons. Those neurons transmit to the output side (their right side), in addition to their nominal value, the error $E$ of the $L$-layer neural network *on the left of layer (L+1)*, multiplying it by at most the maximum synaptic weight from layer L to layer $(L+1)$, $w_m^{(L+1)}$ and the Lipschitz constant K, yielding an error of at most $E(N_{L+1} - f_{L+1})K$.

By the induction hypothesis we have:

$$E \leq C \sum_{l=1}^{L} f_l K^{L-l} \prod_{l'=l+1}^{L+1} (N_{l'} - f'_l) w_m^{(l')}$$

As the output node is linear (not part of the neural network and not performing any non-linear activation function), the errors mentioned in 1 and 2 are added and yield a total error bounded as follows:

$$\|F_{neu}(\mathbf{X}) - F_\lambda(\mathbf{X})\| \leq f_{L+1}w_m^{(L+2)}C + (N_{L+1} - f_{L+1})KE$$

$$\leq C \sum_{l=1}^{L+1} f_l K^{L+1-l} \prod_{l'=l+1}^{L+2} (N_{l'} - f'_l)w_m^{(l')}$$

which is the desired bound for an $(L+1)$-layer network. The equality case follows from considering the inter-occurrence of the equality cases at all the contributing parts, in case no constraint on the network is set to avoid it.

By induction, Theorem 2 is true for any integer $L \geq 1$. ∎

*B. Tight Bound on Neuron Failures*

With the notations used before, in a network of $L$ layers, each layer $l$ containing $N_l$ neurons, we consider $N_{fail} = (f_l)_{l=1}^L$ as the distribution per layer of Byzantine neurons ($f_l$ being the contribution of layer $l$ to $N_{fail}$). Using Theorem 2, and the same reasoning as in the proof of Theorem 1, it is possible to derive a tight bound, not on the total number of failures as in single layer case, but on the failures per layer distribution[10] $N_{fail} = (f_l)_{l=1}^L$.

**Theorem 3.** *Let $F$ be any continuous function mapping $[0,1]^d$ to $[0,1]$, let $\epsilon$ and $\epsilon'$ be any two positive real numbers such that $0 < \epsilon' \leq \epsilon$. Given any $F_{neu}$ that is an L-layer neural $\epsilon'$-approximation of $F$ with $N_l$ neurons per layer $l$, given $N_{fail} = (f_l)_{l=1}^L$ such that $\forall l f_l < N_l$ and*

$$Fep \leq \epsilon - \epsilon'. \tag{10}$$

*Then $F_{neu}$ tolerates the distribution of Byzantine neurons $N_{fail}$. The bound (10) is tight.*

*Proof:* Denote by $F_{fail}$ the output of the network after $N_{fail}$ failures, using Theorem 2 we have: $\|F_{neu}(\mathbf{X}) - F_{fail}(\mathbf{X})\| \leq C \sum_{l=1}^L f_l K^{L-l} \prod_{l'=l+1}^{L+1} (N_{l'} - f'_l)w_m^{(l')}$. Combining this with inequalities 4 and 5 we obtain: $\|F(\mathbf{X}) - F_{fail}(\mathbf{X})\| \leq \epsilon' + C \sum_{l=1}^L f_l K^{L-l} \prod_{l'=l+1}^{L+1} (N_{l'} - f'_l)w_m^{(l')}$. If $N_{fail}$ satisfies inequality 10 then we have $\|F(\mathbf{X}) - F_{fail}(\mathbf{X})\| \leq \epsilon$. This proves the upper bound. Tightness follows the worst case reasoning on the equality and limit cases as done in the proof of Theorem 1. ∎

To better appreciate the message of Theorem 3, one has to bare in mind that the left-hand side of Equation 7 comes from the *forward error propagation* due to the failure distribution $N_{fail}$ (Theorem 2) while the right-hand side is the the maximal error permitted by the over-provision.

Note that, in Theorem 2, small $K$ and small weights reduce the propagating error $Fep$, which translates in Theorem 3 to: the smaller the $K$ and the weights, the easier it is to satisfy the condition with large $f_l$. This sets the basis for the trade-off on tuning $K$ or reducing the weights, as we stated in the introduction and as we discuss in the last section. Note also that in the case of crashes without Byzantine neurons, Assumption 1 is not necessary and $C$ can be replaced by the maximum of the activation function (1 in case of *sigmoid*), which is the maximum value a neuron can send. Note finally that Lemma 1 can also be derived as a limit case of Theorem 3: $N_{fail} \xrightarrow{C \to \infty} 0$.

[10] We use the natural extension of Definition 3 to this generalization from an integer $N_{fail}$ to an $L$-tuple of failures per layer.

*C. The Failure of Synapses*

The following lemma links errors at synapses to errors at neurons. Again we use the convention that layer $L+1$ corresponds to the output node, in addition to the convention that layer 0 corresponds to input nodes.

**Lemma 2.** *In any L-layer neural network, an error of value $\lambda_{ji}^{(l)}$ at the synapse from neuron $i$ of layer $l-1$ to neuron $j$ of layer $l$ is at worst, equivalent to an error at neuron $i$ of value $C.K$.*

*Proof:* Let $l$ be a layer in the neural network, and let $i$ and $j$ be any neurons from $l-1$ and $l$ respectively.

An error of value $\lambda_{ji}^{(l)}$ in the synapse from neuron $i$ to neuron $j$ yields a received sum at neuron $j$, noted $s_{\lambda,j}^{(l)}$ and given by Equation 1 as follows:

$$s_{\lambda,j}^{(l)} = \sum_{k=1, k \neq i}^{N_{l-1}} w_{jk}^{(l)} y_k^{(l-1)} + w_{ji}^{(l)} y_i^{(l-1)} + \lambda_{ji}^{(l)}$$

$$= \sum_{k=1}^{N_{l-1}} w_{jk}^{(l)} y_k^{(l-1)} + \lambda_{ji}^{(l)}$$

Therefore, by K-Lipschitzness of the activation function, the output error of neuron $j$ is bounded as follows:

$$|error| = |\varphi(s_{\lambda,j}^{(l)}) - \varphi(s_j^{(l)})| \leq K.|s_{\lambda,j}^{(l)} - s_j^{(l)}| = K.|\lambda_{ji}^{(l)}|$$

In the worst case the transmitted error $|\lambda_{ji}^{(l)}|$ is equal to $C$ following Assumption 1 and the bound $|error| \leq C.K$ is tight. ∎

**Theorem 4.** *Given $N_{fail} = (f_l)_{l=1}^{L+1}$, the distribution of Byzantine synapses, with $f_l$ being the number of failing ones linking layer $l-1$ to layer $l$: If $C \sum_{l=1}^{L+1} f_l K^{L+1-l} w_m^{(l)} \prod_{l'=l+1}^{L+1} (N_{l'} - f'_l)w_m^{(l')} \leq \epsilon - \epsilon'$ then $F_{neu}$ tolerates the distribution of Byzantine synapses $N_{fail} = (f_l)_{l=1}^{L+1}$. This bound is tight.*

*Proof:* Lemma 2 implies that the failure of a distribution $N_{fail}$ of synapses in an $L$-layer network is equivalent, in the worst case, to the failure of a distribution $N_{fail}$ of neurons in an $L+1$ network. Applying Theorem 3, the result follows. ∎

*D. Reduced Over-provisioning*

Our condition, under which over-provisioned networks can be robust (Theorem 3), concerns networks reaching a precision $\epsilon'$ finer than the one they are required to keep $\epsilon$ (i.e $\epsilon' < \epsilon$). One can wonder how hard it is to reach $\epsilon'$ (i.e, how precise should the over-provisioned network be to tolerate $N_{fail}$). The following corollary establishes the feasibility of building robust networks that can be arbitrarily close to non robust ones ($\epsilon' - \epsilon$ being arbitrarily small).

**Corollary 1.** *Let $N_{fail}$ be any set of $L$ integers as described in IV-B, $\epsilon > 0$ any precision level and $F$ any target function. Then for every $\epsilon' > 0$ such that $\epsilon' < \epsilon$, there exist a neural network approximating $F$ with precision $\epsilon'$ and preserving precision $\epsilon$ under failure distribution $N_{fail}$.*

*Proof:* The existence of a network $\epsilon'$-approximating $F$ is guaranteed by the universal approximation theorem [32] applied to $\epsilon'$ and $F$. Let $w_m^{(l)}$ be the maximal weights at each layer of this network, for the robustness constraint, let $(N_l)_{1 \le l \le L}$ be any set of integers, large enough such that the condition of Theorem 3 is satisfied with $N_{fail}$, $w_m^{(l)}$ and $\epsilon - \epsilon'$. Then, following Theorem 3, this network is an $\epsilon$-approximation of $F$ that tolerates the failure distribution $N_{fail}$. ∎

## V. APPLICATIONS

### A. Reducing Memory Cost

When implementing neural networks in hardware, reducing memory cost typically goes with reducing the precision with which each neuron performs its local computation. However, reducing this local precision impacts accuracy. Recently, experimental results [31] reported interesting trade-offs between cost reduction and accuracy of the output. We provide here the first theoretical result quantifying those trade-offs.

In the case of a neural network containing $L$ layers where the cost reduction implies a maximum error of $\lambda_l$ per layer $l$, the accuracy degradation in the output is bounded by a sum similar to what we give in Theorem 3. This application is not a direct consequence of Theorem 3 but can be more specifically derived from the observations made in the proof of Theorem 2, in which we replace the uniform bound $C$ on the transmission capacity by local bounds $\lambda_l$ per layer. We get the following theorem.

**Theorem 5.** *If in each layer $l$, the implementation induces an error at each neuron $j$ of layer $l$ bounded by $\lambda_l$, then the effect on the output is bounded as follows:*

$$\|F_{neu}(\boldsymbol{X}) - F_\lambda(\boldsymbol{X})\| \le \sum_{l=1}^{L} K^{L-l} \lambda_l \prod_{l'=l}^{L} N_{l'} w_m^{(l'+1)} \quad (11)$$

*where $F_{neu}$ is the nominal neural function, $F_\lambda$ the neural function accounting for the errors $\lambda_l$, $w_m^{(l)} = max(\|w_{ji}^{(l)}\|, (j,i) \in [1, N_l][1, N_{l-1}])$ the maximum norm of the weights of the incoming synapses to layer $l$, and $K$ the Lipschitz coefficient of the activation function. Inequality 11 is tight.*

*Proof:* The proof is similar to that of Theorem 2. We proceed by induction on $L$, the number of layers in a neural network.

**Initiation.** In a single-layer neural network with $N$ neurons, if each neuron $i$ introducing an error $\lambda_i$, and all errors bounded by $C$, then the difference between the nominal output and the output affected by errors is given by $\sum_{i=1}^{N} \lambda_i w_i^{(2)}$ where $w_i^{(2)}$ is the weight from neuron $i$ to the output, therefore the total error is bounded by $NCw_m^{(2)}$ where $w_m^{(2)}$ is the maximum weight from the single layer to the output. This is the base case ($L = 1$) of our induction.

**Induction step.** Let $L$ be an integer such that we have the result of Inequality 11 for every network of $L$ layers. Let $E_L = \sum_{l=1}^{L} K^{L-l} \lambda_l \prod_{l'=l}^{L} N_{l'} w_m^{(l'+1)}$.

Consider now a network of $L+1$ layers. As for the proof of Theorem 2, every neuron $i$ of layer $L+1$ is the output of an $L$ layer network. Neuron $i$ therefore receives a sum affected by an error of at most $E_L$ to which it adds its own error $\lambda_i^{L+1}$, and applies the activation function which multiplies the total error by at most $K$.

The total error at each neuron $i$ is therefore bounded by $KE_l + K\lambda_{L+1}$. Applying the base case at the final output of the network, we bound the final error by $N_{L+1} w_m^{(L+2)} (KE_l + K\lambda_{L+1})$ which is equal to $\sum_{l=1}^{L+1} K^{L-l} \lambda_l \prod_{l'=l}^{L+1} N_{l'} w_m^{(l'+1)}$ and proves the induction. ∎

### B. Boosting Computations

Consider a network where neurons do not have the same reactive speed to inputs, but can be reset instantly to ignore their actual computation. Each time a neuron receives a *sufficient* amount of information from its preceding input layer, it sends a reset to the slow neurons (in the preceding layer) instead of waiting for their values and move on with its own computation, adopting value 0 for the slow neurons. Theorem 3 gives a sense of that *sufficient* amount of information from which we derive the following corollary.

**Corollary 2.** *Following the notation of Theorem 3 in the crash case ($C = 1$ as explained in IV-B), If $F_{neu}$ is an $\epsilon'$-approximation of $F$, then given any family of integers $f_l$ satisfying the conditions of Theorem 3, then each neuron of layer $l$ has to wait only for $N_{l-1} - f_{l-1}$ signals from layer $l - 1$ to send a value to layer $l + 1$, as well as a reset to the missing neurons at layer $l - 1$, while guaranteeing a correct $\epsilon$-approximation of $F$ at the output.*

*Proof:* The corollary is a direct consequence of Theorem 3. ∎

### C. Balancing Robustness and Ease of Learning

Improving the robustness of a neural network can be viewed as minimizing $Fep$ (the right hand term of the inequality in Theorem 2) during the learning scheme.

This would ensure that the neural network has learned the optimal weight distribution and is taking full advantage of the over-provisioning. Clearly, over-provisioning to guarantee $\epsilon'$-accuracy impacts the amount of data needed for learning without over-fitting. This creates a dilemma that somehow resembles the famous bias/variance dilemma [35] in machine learning. In our case, this corresponds to a robustness/ease-of-learning dilemma. The trade-off has two forms we detail below.

**Trade-off on the Lipschitz constant of the activation function** ($K$)**.** Choosing a low value of $K$ leads to satisfying the inequalities of theorems 3 and 4 with high numbers of faults (dependency on $K^{L-l}$). But one should recall that $K$ is an estimate of how sharp the discrimination between inputs at the level of a single neuron is (Figure 2). Therefore, for a network with a low-$K$ activation function, the learning time and the number of necessary neurons can be higher than with a high-K activation function, for the latter is more discriminating.

**Trade-off on synaptic weights.** Like for the Lipschitz-constant $K$, one can note in theorems 3 and 4 (multiplications by the weight) that imposing low weights leaves some room for higher numbers of faults while still satisfying the bound. Achieving this goes through increasing the number of neurons. Intuitively, more neurons are needed to sum to the desired value, if the weights are lower.

## VI. CONCLUDING REMARKS

We established tight bounds relating the output accuracy loss of a neural network to failures of its neurons and synapses. Our bounds are derived from a quantity, $Fep$, the forward error propagation (given in Theorem 2), relating the propagation of imprecision in a neural network to specific parameters of that network, namely, weights, transmission capacity of synapses, coefficient of Lipschitzness of the activation function and number of neurons per layer. The bounds provide a theoretical explanation for some of the cost reduction strategies observed experimentally [31]. Leveraging these bounds, we provided a scheme to boost the synchronization of neural networks and we identified key trade-offs between robustness on the one hand, and learning cost on the other hand.

Whilst our results were established in the context of a feed forward neural network, the underlying methodology (theorems 1, 2 and 3) can be applied to other neural computing models. In the case of the convolutional network model [5], the neurons have a limited receptive field (they are not connected to all other neurons in the adjacent layers like in the feed-forward case), and the weights have a periodic distribution within each layer reducing the size of the set of synapses, which leads in turn to less restrictive bounds (i.e tolerating larger amount of failures). More precisely, the maximal weight constraint $w_m^{(l)}$ appearing in theorems 2 and 3 will run only[11] on the $R(l)$-different values of the weights from layer $l-1$ to layer $l$, $R(l)$ being the size of the receptive field of layer $l$ (i.e to how many neurons of layer $l-1$ each neuron of layer $l$ is connected).

More generally, we believe that the methodology that led to our results can be applied to any set of distributed processes achieving a global computation while putting weights on the values of each other and (unlike classical settings in distributed computing [14]) are not performing the same computation in each node and do not have to agree on values, as long as a similar condition to Assumption 1 applies (bounded channel capacity).

To conclude, it is important to note that our results were established independently of the chosen learning scheme. An appealing research direction is to consider a specific learning scheme taking the forward error propagation as an additional minimization target which would reduce the impacts of failures. To our knowledge, there has been one single attempt to theoretically formulate such an optimization problem [36], but it only minimizes the effect of the crash of a *single* neuron. Our bounds help formulate the distributed optimization problem for a multiple neurons and synapses, which opens the question of the computational cost of building a neural network that achieves a given robustness constraint with such a learning scheme.

*Acknowledgment*

The authors would like to thank Julien Stainer for useful bibliographic pointers and comments on the manuscript.

This work has been supported in part by the Swiss National Science Foundation, Grant 200021 169588 / TARBDA.

---

[11] With the notation of theorems 2 and 3, let us consider a convolutional network as a multilayer feed forward neural network such as for each synapse connecting a neuron $i$ in layer $l$ to a neuron $j$ in layer $l-1$ not belonging to the receptive field of $i$, we have $w_{ji}^{(l)} = 0$. Together with the weight sharing property of convolutional networks this leads to the equality between the maximal weight in absolute value over a layer $w_m^{(l)}$ and the maximum weight in absolute value over a single receptive field.